\documentclass{article}

\usepackage{arxiv}

\usepackage[utf8]{inputenc}             
\usepackage[T1]{fontenc}                 
\usepackage[colorlinks]{hyperref}       
\usepackage{url}                        
\usepackage{booktabs}                   
\usepackage{amsfonts}                   
\usepackage{nicefrac}                   
\usepackage{microtype}                  
\usepackage{lipsum}
\usepackage{graphicx}

\usepackage{amsmath}
\usepackage{subcaption}
\usepackage{adjustbox}
\usepackage{multirow}
\usepackage{authblk}

\graphicspath{ {./figs/} }

\title{Swish-T : Enhancing Swish Activation with Tanh Bias for Improved Neural Network Performance}

\author{
  Youngmin Seo, Jinha Kim, Unsang Park\thanks{Corresponding Author: unsangpark@sogang.ac.kr} \\
  Sogang University \\
  \{ymin98, jhkmo510, unsangpark\}@sogang.ac.kr
}
\date{}

\begin{document}

\maketitle

\begin{abstract}
    We propose the Swish-T family, an enhancement of the existing non-monotonic activation function Swish. Swish-T is defined by adding a Tanh bias to the original Swish function. This modification creates a family of Swish-T variants, each designed to excel in different tasks, showcasing specific advantages depending on the application context. The Tanh bias allows for broader acceptance of negative values during initial training stages, offering a smoother non-monotonic curve than the original Swish. We ultimately propose the Swish-T$_{\textbf{C}}$ function, while Swish-T and Swish-T$_{\textbf{B}}$, byproducts of Swish-T$_{\textbf{C}}$, also demonstrate satisfactory performance. Furthermore, our ablation study shows that using Swish-T$_{\textbf{C}}$ as a non-parametric function can still achieve high performance. The superiority of the Swish-T family has been empirically demonstrated across various models and benchmark datasets, including MNIST, Fashion MNIST, SVHN, CIFAR-10, and CIFAR-100. The code is publicly available at ``https://github.com/ictseoyoungmin/Swish-T-pytorch''.
\end{abstract}

\keywords{Activation Function \and Deep Learning \and Neural Network}

\section{Introduction}

    Activation functions are crucial in deep learning for introducing non-linearity, which is essential for modeling complex patterns. Early activation functions like Sigmoid and Tanh enabled smooth transitions but were susceptible to the vanishing gradient problem. The introduction of the Rectified Linear Unit (ReLU)\cite{hahnloser2000digital,jarrett2009best,nair2010rectified} marked a significant advancement due to its simplicity and efficiency, enhancing gradient flow and accelerating learning. However, despite its benefits, ReLU can suffer from the "dying ReLU" problem, where neurons become inactive and cease to function across a range of inputs.

\begin{figure*}[ht]
        \hfill 
        \begin{subfigure}{\textwidth}
            \centering
            \includegraphics[width=\textwidth]{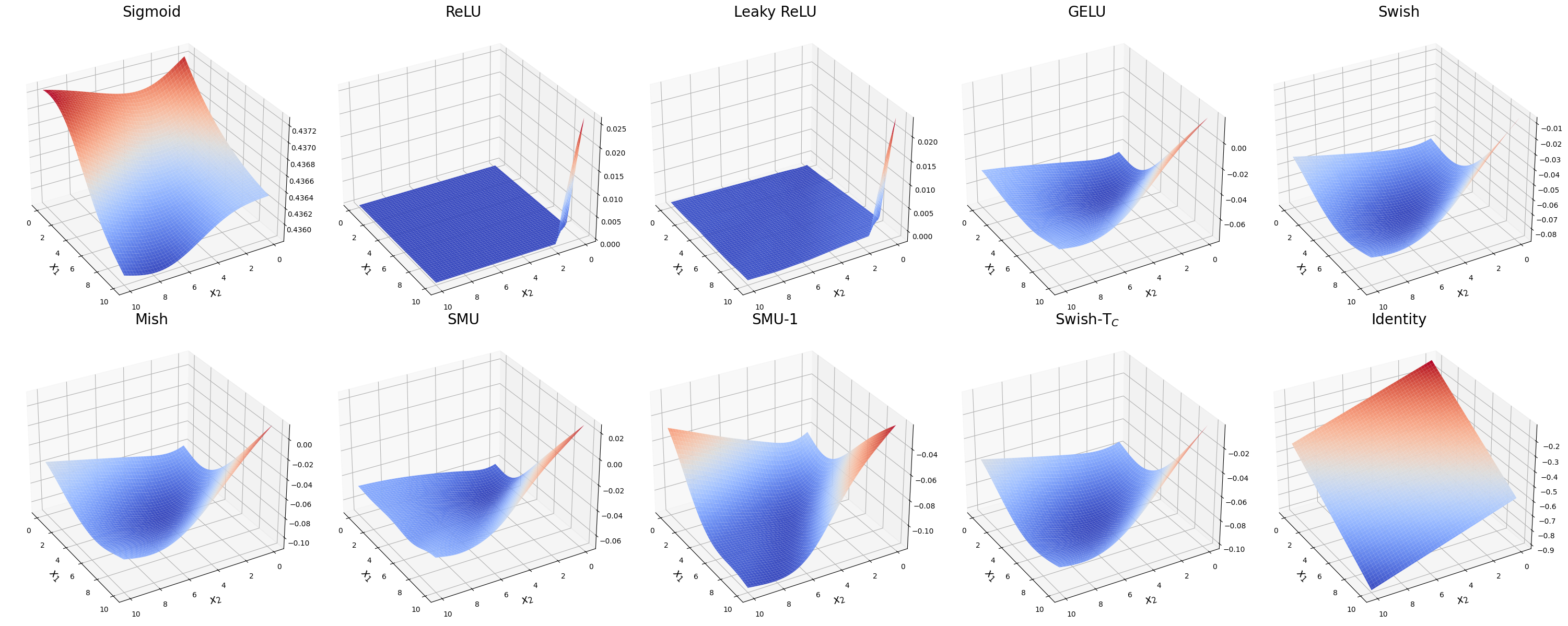}
        \end{subfigure}
        \vspace{0.2cm}
        \caption{Comparison of Various Activation Functions including Sigmoid, ReLU, Leaky ReLU, GELU, Swish, Mish, SMU, SMU-1, Swish-T$_{\textbf{C}}$, and Identity in the Output Landscape of a 3-Layer Neural Network. All the networks' weights are randomly initialized, and no training has been performed, showcasing the initial output patterns induced by each activation function.}
        \label{fig:landscape}
    \end{figure*}

    To analyze the impact of various activation functions, we visualize the output landscape of a 3-layer neural network with randomly initialized weights, without training, as shown in Fig.~\ref{fig:landscape}. Fig.~\ref{fig:landscape} highlights even before learning begins. The "dying ReLU" issue can be considered as incurring from the flat regions where ReLU neurons output zero, resulting in large inactive areas in the landscape.
    
    To overcome the limitations of ReLU, variants such as Leaky ReLU~\cite{lrelu} and Parametric ReLU (PReLU)~\cite{he2015delving} have been developed. These modifications address the "dying ReLU" issue by allowing a small, non-zero gradient when the unit is inactive, thus maintaining learning capabilities during backpropagation. Nonetheless, these functions introduce a new challenge as they are non-differentiable at zero, leading to a discontinuity in the derivative. This non-differentiability can complicate optimization, as the precise point where this occurs may result in unpredictable behaviors in gradient-based learning methods.
    
    Further advancements have been made with functions like GELU~\cite{hendrycks2023gaussian}, Swish~\cite{ramachandran2017searching}, ACONs~\cite{Ma_2021}, Pserf~\cite{Biswas_2022}, ErfAct~\cite{Biswas_2022}, and Mish~\cite{misra2020mish}, which are smooth approximations of ReLU, thereby alleviating some of its issues. Notably, Mish, inspired by Swish, and ACONs, introducing a dynamic choice between linear and non-linear modes, enhances neural network's adaptability and generalizes Swish as a special case of ACON-A~\cite{Ma_2021}. PSerf, as a non-monotonic, smooth and parametric activation function, demonstrates significant performance improvements over traditional activations like ReLU and Swish. SMU~\cite{9878772} has shown significant performance improvements by providing smooth approximations to the Maxout~\cite{goodfellow2013maxout} family functions, including ReLU and Leaky ReLU.
    
    Among the successors to ReLU, Swish has emerged as a particularly effective function. As a result, various Swish-based functions such as E-Swish~\cite{alcaide2018eswish}, P-Swish~\cite{9301059}, and Soft-Clipping Swish~\cite{9465622} have been proposed. These functions improve performance by adding parameters to the original Swish or by combining it with other functions, enhancing their capability to model complex patterns.

    The contribution of this paper is summarized as follows:

    \begin{itemize}
    \item \textbf{Introduction of Swish-T Family}:
     The Swish-T family enhances the existing Swish activation function by adding a Tanh bias, allowing broader acceptance of negative values and providing a smoother non-monotonic curve.

    \item \textbf{Variants of Swish-T}:
    Swish-T$_{\textbf{A}}$, Swish-T$_{\textbf{B}}$, and Swish-T$_{\textbf{C}}$ are variants designed for improved computational efficiency, adaptability, and stability, respectively.

    \item \textbf{Empirical Validation}:
     Extensive experiments demonstrate the superior performance of the Swish-T family across various datasets (MNIST, Fashion MNIST, SVHN, CIFAR-10, CIFAR-100) and architectures (ResNet-18, ShuffleNetV2, SENet-18, EfficientNetB0, MobileNetV2, DenseNet-121).
    \end{itemize}


\section{Related work and Motivation}

    Recent studies, including those by SMU, ACONs, and Pserf, have focused on generalizing activation functions such as ReLU, Leaky ReLU, and PReLU through smooth approximations while incorporating trainable parameters. These studies suggest that Swish, despite its simple definition, demonstrates powerful performance and holds the potential for further evolution into more advanced forms.
    
    The Swish function, defined as \( y = x \cdot \text{Sigmoid}(\beta x) \), has emerged as a powerful activation function, demonstrating superior performance in various deep learning tasks such as image classification~\cite{math7121170}, object detection~\cite{Tan_2020,farheen2020skin}, and Natural Language Processing (NLP)~\cite{Eger_2018}. Discovered through Neural Architecture Search (NAS), Swish is characterized by its non-monotonicity and smoothness, which allow it to maintain small negative weights, enhancing model training and performance stability.
    
    Swish offers several advantages:
    \begin{itemize}
        \item \textbf{Smooth Nonlinearity}: Swish is a smooth, non-monotonic function, which improves information propagation in deep networks.
        \item \textbf{Self-gating}: Swish includes a self-gating mechanism that adjusts the output scale based on the input value, allowing for a more flexible function form.
        \item \textbf{Training Efficiency and Performance}: Empirical results indicate that Swish outperforms ReLU in deep models and large datasets, handling gradient vanishing and saturation problems more effectively.
    \end{itemize}
    
    Building on these strengths, we introduce a new concept by incorporating bias terms directly within the activation function, inspired by the significant role of bias in neural network layers~\cite{pmlr-v97-wang19p}. We hypothesize that incorporating bias terms into the activation function itself can provide more sophisticated control of the activation threshold, enabling a more nuanced adaptation to the input function. This hypothesis is based on the fact that bias in an existing layer better adjusts the input signal for a desired output range, thereby improving the learning ability of the network.
    
    Our research proposes the Swish-T family, which incorporates a parameterized bias term into the Swish function. This novel variant aims to optimize performance further and enhance adaptability across various architectures and applications. By integrating the bias term, we seek to improve the learning dynamics and overall network performance.
    
\section{Swish-T}

    \begin{figure*}[t]
        \hfill 
        \begin{subfigure}{\textwidth}
            \centering
            \includegraphics[width=\textwidth]{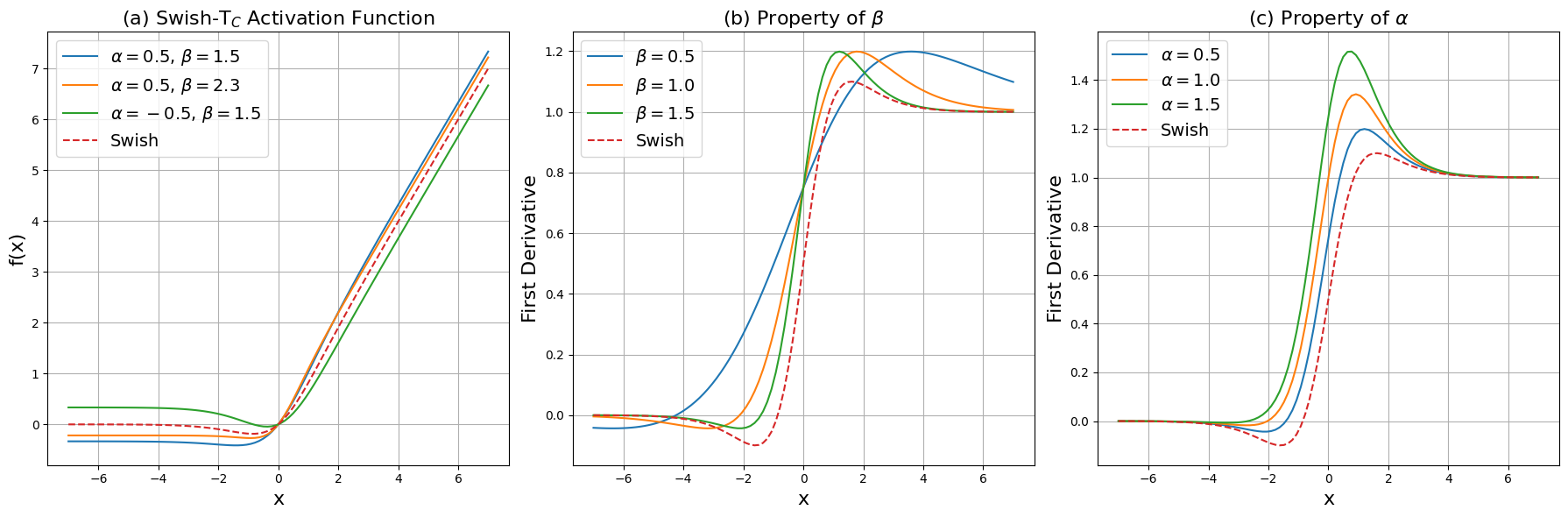}
        \end{subfigure}
        \vspace{0.1cm}
        \caption{Swish-T$_{\textbf{C}}$, Swish activation function and first derivatives. (a) Swish-T$_{\textbf{C}}$ activation function with fixed alpha and beta. (b) The first derivatives with fixed alpha$=$0.5 and different betas. Beta controls how quickly the first derivative reaches the upper/lower asymptotes. (c) Alpha determines the upper/lower bounds of the first derivative.}
        \label{fig:property}
    \end{figure*}
    
    We present the Swish with Tanh bias function (Swish-T) family. To complement Swish, we conducted preliminary research to find an appropriate bias function. The criteria for the bias function are as follows:
    
    \begin{itemize}
        \item \textbf{Zero-centered}: This ensures that the function maintains the zero-centered characteristic of the original Swish function.
        \item \textbf{Non-linearity}: The bias function should be nonlinear to effectively adjust the bias based on the value of \(x\).
        \item \textbf{Bounded}: It should be bounded by a scaling parameter \(\alpha\) to ensure it does not disrupt the properties of the original Swish function.
    \end{itemize}
    
    Among the combinations of $\alpha x^2$, $\alpha \tanh(x)$, and $\alpha \tanh(\beta x)$, the configurations Swish + $\alpha \tanh(x)$ and Swish-1 + $\alpha \tanh(\beta x)$ showed competitive performance. Specifically, Swish + $\alpha \tanh(x)$ achieved a top-1 accuracy of 79.02 on the CIFAR-100 dataset~\cite{cifar} using the ResNet-18 model~\cite{resnet}, leading to the derivation of the Swish-T function family. In comparison, Swish + $\alpha x^2$ and Swish-1 + $\alpha \tanh(\beta x)$ obtained top-1 accuracies of 77.31 and 78.22, respectively.
    
    \subsection{Swish-T Family}
    The basic design of Swish-T is adding \(\alpha \tanh(x)\) to the original Swish function, expressed as:
    \begin{equation}
        \begin{aligned}
            f(x; \beta, \alpha) &= x \sigma(\beta x) + \alpha \tanh(x) \\
            &= x \sigma(\beta x) + \alpha (2\sigma(2x) - 1) \\
            &= x \sigma(\beta x) + 2 \alpha \sigma(2x) - \alpha 
        \end{aligned}
        \label{swisht}
    \end{equation}
    
    Here, \(\sigma(x)\) denotes the sigmoid function, \(\beta\) is a trainable parameter, and \(\alpha\) is a hyper-parameter used to scale the tanh function from \((-1, 1)\) to \((-\left| \alpha \right| , \left|\alpha \right|)\). In our experiments, the Swish-T family was initialized with \(\beta = 1.0\) and \(\alpha = 0.1\), and \(\beta\) was updated during training.
    
    Swish-T converges to \(\alpha\) in the negative region. The bias increases or decreases until bounded by the tanh function and is determined by \(x\). When \(x = 0\), the tanh output is 0, maintaining the zero-centered property of the original Swish function.
    
    Eq.~\ref{swisht} cannot be simplified because the coefficients of \(x\) inside the sigmoid function differ. Simplifying the equation could improve the forward and backward pass speeds in deep learning frameworks like TensorFlow \cite{tensorflow} and PyTorch \cite{pytorch}. To enhance speed while maintaining or improving performance, we designed the Swish-T family, including Swish-T$_{\textbf{A}}$, Swish-T$_{\textbf{B}}$, and Swish-T$_{\textbf{C}}$.
    
    Swish-T$_{A}$ simplifies Eq.~\ref{swisht} by unifying the coefficients of \(x\) to 1, improving speed:
    \begin{equation}
        \begin{aligned}
            f_A(x; \alpha) &= x \sigma(x) + 2 \alpha \sigma(x) - \alpha \\
            &= \sigma(x)(x + 2 \alpha) - \alpha 
        \end{aligned}
        \label{swisht_a}
    \end{equation}
    
    This function is equivalent to Swish-1(\(x\)) + \(\alpha \tanh(x/2)\) and has no trainable parameters. This transformation is advantageous in scenarios where computational efficiency is crucial.
    
    Swish-T$_{\textbf{B}}$ reintroduces a trainable parameter to Eq.~\ref{swisht_a}:
    \begin{equation}
        \begin{aligned}
            f_B(x; \beta, \alpha) &= \sigma(\beta x)(x + 2 \alpha) - \alpha 
        \end{aligned}
        \label{swisht_b}
    \end{equation}
    
    Rewriting Eq.~\ref{swisht_b} as \(x \sigma(\beta x) + \alpha \tanh(\beta/2 x)\) shows that the bias depends on both \(x\) and \(\beta\).
    
    Swish-T$_\textbf{C}$ ensures symmetry when \(\beta\) changes sign:
    \begin{equation}
        \begin{aligned}
            f_C(x; \beta, \alpha) &= \sigma(\beta x)(x + \frac{2 \alpha}{\beta}) - \frac{\alpha}{\beta}, \quad \text{where \,} \beta \neq 0 
        \end{aligned}
        \label{swisht_c}
    \end{equation}
    
    Simplifying Eq.~\ref{swisht_c} shows that the maximum bias coefficient is \(\alpha / \beta\). As \(\beta\) approaches infinity, the bias converges to 0, making it act like a smooth ReLU. When \(\beta\) approaches 0, the function approaches 0. With \(\alpha = 1.0\) and \(\beta = 1e^{-4}\), this function behaves as the Identity function. Adjusting \(\alpha\) changes the slope, requiring precise tuning for exact Identity function behavior. This transformation is useful for applications requiring stable performance across a wide range of inputs.
    
    Each function variant exhibits unique convergence characteristics. Swish-T$_{\textbf{A}}$ simplifies to enhance computational efficiency. Swish-T$_{\textbf{B}}$ dynamically adapts the bias with \(\beta\), suitable for tasks with diverse input characteristics. Swish-T$_{\textbf{C}}$ ensures symmetry when \(\beta\) changes sign, maintaining stable performance across various input ranges. The bounded nature of the tanh function ensures that the bias remains within a manageable range, preventing extreme changes that could destabilize training. This is particularly evident in (Swish-T, Swish-T$_{\textbf{A}}$, Swish-T$_{\textbf{B}}$) and Swish-T$_{\textbf{C}}$, where the bias converges to \(\pm \alpha\) and  \(\pm \alpha / \beta\), respectively.


\subsection{Gradient Computation for Swish-T Family}
    In the context of training neural networks, the efficiency of backpropagation\cite{backpropagation1974} is crucial. This section derives the gradient expressions for Swish-T and its variants, highlighting their impact on backpropagation.

    The gradient of the Swish-T activation function can be expressed as:
    \begin{equation}
         \begin{aligned}
        \frac{d}{dx} f = \sigma(\beta x) + \beta x \sigma(\beta x) (1 - \sigma(\beta x)) + 4 \alpha \sigma(2x) (1 - \sigma(2x)), \\
        \text{where}\, \frac{d}{dx} \sigma(x) = \sigma(x)(1 - \sigma(x))
        \end{aligned}
        \label{d_swisht}
    \end{equation}
    
    Among the Swish-T family, Swish-T$_{\textbf{A}}$ shows the fastest learning speed. Its derivative is given by:
    \begin{equation}
        \begin{aligned}
        \frac{d}{dx} f_A &= \sigma(x)(1 - \sigma(x)(x + 2\alpha) + \sigma(x)) \\
        &= \sigma(x)((1 - \sigma(x)(x + 2\alpha)) + 1) \\
        &= \sigma(x)(x + \alpha + 1 - (x\sigma(x) + 2\alpha\sigma(x) - \alpha)) \\ 
        &= \sigma(x)(x + \alpha + 1 - f_A(x)) 
        \end{aligned}
        \label{d_swisht_a}
    \end{equation}
    
    This expression, along with Eq.~\ref{d_swisht_a}, shows that Swish-T$_{\textbf{A}}$ can be differentiated efficiently, making it suitable for backpropagation in terms of both memory and speed. Similarly, the gradients for Swish-T$_{\textbf{B}}$ and Swish-T$_{\textbf{C}}$ include their own expressions and are derived as follows:
    
    For Swish-T$_{\textbf{B}}$:
    \begin{equation}
        \begin{aligned}
        \frac{d}{dx} f_B &= \beta \sigma(\beta x)(1 - \sigma(\beta x)(x + 2\alpha) + \sigma(\beta x)) \\
        &= \sigma(\beta x)(\beta (x + \alpha - f_B(x)) + 1)
        \end{aligned}
        \label{d_swisht_b}
    \end{equation} 
    
    Eq.~\ref{d_swisht_b} describes the derivative of the Swish-T$_{\textbf{B}}$ activation function. The term \(\beta \sigma(\beta x)\) represents the impact of the activation function on the gradient, with the factor \(\beta\) scaling the input \(x\). The expression \((1 - \sigma(\beta x)(x + 2\alpha) + \sigma(\beta x))\) captures the adjustment made by the sigmoid function, \(\sigma(\beta x)\), and the constants \(\alpha\) and \(\beta\). This formulation ensures a controlled gradient flow, balancing the input \(x\) and the influence of the activation function.
    
    For Swish-T$_{\textbf{C}}$:
    \begin{equation}
        \begin{aligned}
        \frac{d}{dx} f_C &= \beta \sigma(\beta x)(1 - \sigma(\beta x)(x + 2\alpha / \beta) + \sigma(\beta x)) \\
        &= \sigma(\beta x)(\beta (x - f_C(x))+ \alpha + 1)
        \end{aligned}
        \label{d_swisht_c}
    \end{equation} 
    
    Eq.~\ref{d_swisht_c} describes the derivative of the Swish-T$_{\textbf{C}}$ activation function. Unlike Swish-T$_{\textbf{B}}$, this formulation avoids the direct multiplication of \(\alpha\) and \(\beta\), which helps prevent situations where the maximum gradient could rapidly increase. This characteristic is evident in Fig.~\ref{fig:property}, where the stability of the gradient flow is demonstrated. The term \(\beta \sigma(\beta x)\) remains, similar to Swish-T$_{\textbf{B}}$, but the expression \((1 - \sigma(\beta x)(x + 2\alpha / \beta) + \sigma(\beta x))\) shows a different interaction between the input \(x\) and the constants \(\alpha\) and \(\beta\). This leads to a more stable gradient behavior, as highlighted by the addition of \(\alpha + 1\) in the final term.
    
    These derivatives show that the Swish-T family of activation functions can be differentiated efficiently, making them suitable for backpropagation. The variations in their formulations offer different advantages in terms of learning speed and gradient stability, as illustrated in the respective equations and supported by empirical evidence.

    Below are the equations for the gradient of the Swish-T family activation functions with respect to \(\beta\):
    
    \begin{equation}
        \begin{aligned}
        \frac{d}{d\beta} f &= x^2 \cdot \sigma(\beta x) (1 - \sigma(\beta x))
        \end{aligned}
        \label{d_beta_swisht}
    \end{equation}
    \begin{equation}
        \begin{aligned}
        \frac{d}{d\beta} f_B &= x \cdot (x + 2\alpha) \cdot \sigma(\beta x) (1 - \sigma(\beta x))
        \label{d_beta_swisht_b}
        \end{aligned}
    \end{equation}
    \begin{equation}
        \begin{aligned}
        \frac{d}{d\beta} f_C &= x \cdot \left( x + \frac{2\alpha}{\beta} \right) \sigma(\beta x) (1 - \sigma(\beta x)) - \frac{2\alpha \sigma(\beta x)}{\beta^2} + \frac{\alpha}{\beta^2}
        \label{d_beta_swisht_c}
        \end{aligned}
    \end{equation}

    In the process of backpropagation, trainable parameters are updated through the gradient of the activation function, specifically the value obtained by differentiating with respect to the parameter in question.

    For the Swish-T activation function, the gradient of the parameter \( \beta \) is given by Eq.~\ref{d_beta_swisht}. This expression shows that the gradient depends on the input \( x \), the sigmoid function \( \sigma(\beta x) \), and its derivative. The \( x^2 \) term indicates that larger inputs have a greater influence on the gradient, leading to faster updates of the \( \beta \) parameter during training.
    
    For the Swish-T$_{\textbf{B}}$ variant, the gradient of \( \beta \) is given by Eq.~\ref{d_beta_swisht_b}. The additional term \( (x + 2\alpha) \) modifies the influence of the input \( x \), suggesting that the parameter \( \alpha \) also plays a role in forming the gradient. This interaction between \( \alpha \) and \( \beta \) allows for fine-tuning the behavior of the activation function, providing more precise control over the learning process.
    
    For the Swish-T$_{\textbf{C}}$ variant, the gradient of \( \beta \) is given by Eq.~\ref{d_beta_swisht_c}. The division by \( \beta^2 \) or \( \beta \) indicates that the magnitude of the gradient is inversely proportional to \( \beta \), which helps prevent excessively large updates during training.
    
    Overall, the effect of \( \beta \) for the Swish-T function emphasizes the complex dependency between the input \( x \), the trainable parameter \( \beta \), and constants such as \( \alpha \). Carefully adjusting \( \beta \) can enhance the efficiency of backpropagation, leading to faster convergence and improved learning outcomes for the neural network.
    
    
\section{Experiments and Results}
    
    We apply both non-parametric and parametric functions uniformly across all activation operations in the network. Consequently, the trainable parameters of our functions and comparison functions are applied globally and updated accordingly. This approach allows us to identify a single optimized function for various architectures. The experiments primarily focus on image classification tasks, with the potential to extend to other tasks such as semantic segmentation and object detection to explore generalization capabilities.
    
    To evaluate the performance of our functions, we compare them against widely used activation functions known for their high performance: ReLU \cite{hahnloser2000digital,jarrett2009best,nair2010rectified}, GELU \cite{hendrycks2023gaussian}, Swish \cite{ramachandran2017searching}, SiLU (Swish-1) \cite{Elfwing_2018}, and Mish \cite{misra2020mish}, as well as the recently introduced, high-performing SMU \cite{9878772} and SMU-1. The \(\beta\) parameter in the Swish-T family is initialized to 1.0, the same as Swish, while the hyperparameter \(\alpha\) is set to 0.1 across all experiments. The trainable parameter \(\mu\) in SMU \cite{9878772} and SMU-1 \cite{9878772} is initialized to 1.0. For SMU, the hyperparameter \(\alpha\) is set to 0.0, while for SMU-1, \(\alpha\) is set to 0.25. All trainable parameters of the activation functions are updated using the backpropagation algorithm \cite{backpropagation1988}. Experiments are conducted on an NVIDIA 2080Ti GPU, with PyTorch \cite{pytorch} version 2.0.1.

\subsection{MNIST, Fashion MNIST and SVHN}
    In this section, we evaluate the performance of various activation functions on three popular image classification datasets: MNIST \cite{mnist}, Fashion MNIST \cite{fashion}, and SVHN \cite{svhn}. The MNIST dataset contains 70,000 grayscale images of handwritten digits (0-9), each of size $28 \times 28$ pixels, widely used for training and testing image processing systems. Fashion MNIST, a dataset similar in structure to MNIST, consists of 70,000 grayscale images of $28 \times 28$ pixels depicting various types of clothing items, offering a more complex alternative to digit classification. The SVHN (Street View House Numbers) dataset includes 600,000 color images of house numbers captured from Google Street View, with each image containing digits in a natural scene, providing a challenging dataset due to its real-world variability. The experiments are conducted using the LeNet \cite{lenet} architecture, a well-known convolutional neural network model designed for handwritten digit recognition. The optimizer used was Stochastic Gradient Descent (SGD) \cite{sgd1,sgd2} with a learning rate of 0.01, momentum of 0.9, and weight decay of 5e-4. The training was carried out for 100 epochs, with the learning rate adjusted using the cosine annealing \cite{coslr} scheduler. Each experiment utilized a batch size of 128. We used standard data augmentation techniques such as random affine transform. Each training process was repeated 10 times. The results can be found in Table~\ref{tab:mnist_results}.

    \begin{table}[h]
    \centering
    \begin{adjustbox}{max width=\dimexpr\textwidth\relax}
        \begin{tabular}{@{}lccc@{}}
        \toprule
        \textbf{Activation Function} & \textbf{MNIST} & \textbf{Fashion MNIST} & \textbf{SVHN} \\ \midrule
        ReLU & 99.34 $\pm$ 0.028 & 89.89 $\pm$ 0.205 & 90.59 $\pm$ 0.203 \\
        GELU & 99.36 $\pm$ 0.034 & 89.84 $\pm$ 0.230 & 90.79 $\pm$ 0.117 \\
        SiLU & 99.31 $\pm$ 0.022 & 89.66 $\pm$ 0.084 & 90.61 $\pm$ 0.099 \\
        Swish & 99.31 $\pm$ 0.010 & 89.80 $\pm$ 0.142 & 90.73 $\pm$ 0.101 \\
        Mish & 99.32 $\pm$ 0.013 & 89.80 $\pm$ 0.151 & 90.65 $\pm$ 0.116 \\
        SMU & 99.30 $\pm$ 0.022 & 89.91 $\pm$ 0.058 & 90.74 $\pm$ 0.230 \\
        SMU-1 & 99.35 $\pm$ 0.039 & 89.82 $\pm$ 0.235 & 90.72 $\pm$ 0.146 \\ \midrule
        Swish-T & 99.34 $\pm$ 0.033 & 89.85 $\pm$ 0.059 & \textbf{90.99} $\pm$ 0.199 \\
        Swish-T$_{\textbf{B}}$ & \textbf{99.37} $\pm$ 0.023 & 89.94 $\pm$ 0.189 & 90.80 $\pm$ 0.096 \\
        Swish-T$_{\textbf{C}}$ & 99.31 $\pm$ 0.029 & \textbf{90.03} $\pm$ 0.099 & 90.80 $\pm$ 0.206 \\ \bottomrule
        \end{tabular}
        \end{adjustbox}
        \vspace{0.2cm}
        \caption{Comparison of various activation functions across the MNIST, Fashion MNIST, and SVHN datasets using the LeNet architecture. The table shows the mean Top-1 test accuracy and standard deviation of 10 runs.}
        \label{tab:mnist_results}
    \end{table}
        
\subsection{CIFAR}
    
    In this section, we report the results on the popular image classification benchmark datasets CIFAR-10 \cite{cifar} and CIFAR-100 \cite{cifar}. These datasets consist of 60,000 color images of 10 and 100 classes, respectively, with each class containing 6,000 images of $32 \times 32$ pixel size. CIFAR-10 contains 50,000 training images and 10,000 test images, while CIFAR-100 has the same number of training and test images of 100 classes.
  
    \begin{table}[h]
        \centering
        \hfill
        \begin{adjustbox}{max width=\dimexpr\textwidth\relax}
            \begin{tabular}{@{}lccccccc@{}}
            \toprule
            \textbf{Activation Function} & \textbf{RN-18} & \textbf{SF-V2 (1.x)} & \textbf{SF-V2 (2.x)} & \textbf{SENet-18} & \textbf{EN-B0} & \textbf{MN-V2} & \textbf{DN-121} \\ \midrule
            \#Params & 11.1M & 1.4M & 5.5M & 11.2M & 3.6M & 2.3M & 1.0M  \\ \midrule
            ReLU & 95.54 $\pm$ 0.08 & 91.90 $\pm$ 0.24 & 91.93 $\pm$ 0.21 & 95.25 $\pm$ 0.32 & 90.73 $\pm$ 0.13 & 92.63 $\pm$ 0.19 & 94.82 $\pm$ 0.16 \\
            GELU & 94.79 $\pm$ 0.16 & 93.06 $\pm$ 0.13 & 93.11 $\pm$ 0.26 & 94.68 $\pm$ 0.10 & 92.14 $\pm$ 0.15 & 93.72 $\pm$ 0.17 & 93.89 $\pm$ 0.21 \\
            SiLU & 94.22 $\pm$ 0.17 & 92.03 $\pm$ 0.11 & 92.34 $\pm$ 0.19 & 93.99 $\pm$ 0.04 & 91.19 $\pm$ 0.15 & 92.59 $\pm 0.16$ & 92.75 $\pm$ 0.17 \\
            Swish & 95.49 $\pm$ 0.07 & 94.06 $\pm$ 0.15 & 94.01 $\pm$ 0.21 & 95.41 $\pm$ 0.14 & \textbf{93.46} $\pm$ 0.13 & 95.08 $\pm$ 0.12 & 94.89 $\pm$ 0.11 \\
            Mish & 94.36 $\pm$ 0.12 & 92.34 $\pm$ 0.14 & 92.41 $\pm$ 0.08 & 94.09 $\pm$ 0.13 & 91.58 $\pm$ 0.14 & 92.81 $\pm$ 0.21 & 92.78 $\pm$ 0.19 \\
            SMU & \textbf{95.58} $\pm$ 0.03 & 94.00 $\pm$ 0.17 & 94.01 $\pm$ 0.12 & 95.48 $\pm$ 0.07 & 93.36 $\pm$ 0.10 & \textbf{95.12} $\pm$ 0.11 & \textbf{94.95} $\pm$ 0.20 \\ 
            SMU-1 & 95.12 $\pm$ 0.16 & 93.94 $\pm$ 0.19 & 93.93 $\pm$ 0.14 & 94.87 $\pm$ 0.15 & 92.98 $\pm$ 0.15 & 94.66 $\pm$ 0.13 & 94.42 $\pm$ 0.18 \\ \midrule
            Swish-T & 95.53 $\pm$ 0.14 & 94.15 $\pm$ 0.14 & 94.08 $\pm$ 0.21 & 95.40 $\pm$ 0.13 & 93.22 $\pm$ 0.13 & 94.89 $\pm$ 0.12 & 94.88 $\pm$ 0.18 \\
            Swish-T$_{\textbf{B}}$ & 95.17 $\pm$ 0.21 & 93.99 $\pm$ 0.18 & 93.91 $\pm$ 0.14 & 95.32 $\pm$ 0.04 & 93.04 $\pm$ 0.28 & 94.72 $\pm$ 0.20 & 94.71 $\pm$ 0.14 \\
            Swish-T$_{\textbf{C}}$ & 95.29 $\pm$ 0.03 & \textbf{94.26} $\pm$ 0.08 & \textbf{94.27} $\pm$ 0.14 & \textbf{95.50} $\pm$ 0.10 & 93.28 $\pm$ 0.19 & 94.97 $\pm$ 0.14 & \textbf{94.95} $\pm$ 0.09 \\ \bottomrule
            \end{tabular}
        \end{adjustbox}
        \vspace{0.2cm}
        \caption{Comparison of the Swish-T family and various activation functions across different architectures on the CIFAR-10 dataset. The table shows the mean Top-1 accuracy and standard deviation of 5 runs.}
        \label{tab:cifar10}
    \end{table}

    The models used for evaluation include ResNet-18 (RN-18) \cite{resnet}, ShuffleNetV2 (SF-V2)(1.x, 2.x) \cite{shufflev2}, SENet-18 \cite{senet}, EfficientNetB0 (EN-B0) \cite{efficientnet}, MobileNetV2 (MN-V2) \cite{mobilenetv2}, and DenseNet-121 (DN-121) \cite{densenet}. For all models, the batch size is set to 128, and the learning rate is initialized to 0.1. The learning rate is adjusted using a cosine annealing \cite{coslr} scheduler, which gradually decreases the learning rate from the initial value to zero over the training period. We use the stochastic gradient descent (SGD) \cite{sgd1,sgd2} optimizer with a momentum of 0.9 and a weight decay of 5e-4. The models are trained for a total of 200 epochs. We used standard data augmentation techniques such as random crop and horizontal flip.
    
    \begin{table}[h]
    \centering
    \begin{adjustbox}{max width=\textwidth}
    \begin{tabular}{@{}lcccccccccccc@{}}
    \toprule
        \textbf{Activation Function} & \textbf{RN-18} & \textbf{SF-V2 (1.x)} & \textbf{SF-V2 (2.x)} & \textbf{SENet-18} & \textbf{EN-B0} & \textbf{MN-V2} & \textbf{DN-121} \\ \midrule
        \#Params & 11.1M & 1.4M & 5.5M & 11.2M & 3.6M & 2.3M & 1.0M \\ \midrule
        ReLU & 78.46 $\pm$ 0.15 & 71.70 $\pm$ 0.41 & 71.92 $\pm$ 0.45 & 77.85 $\pm$ 0.28 & 67.71 $\pm$ 0.49 & 72.65 $\pm$ 0.24 & 76.55 $\pm$ 0.39 \\
        GELU & 78.08 $\pm$ 0.26 & 75.04 $\pm$ 0.21 & 74.83 $\pm$ 0.17 & 76.92 $\pm$ 0.14 & 71.93 $\pm$ 0.34 & 75.83 $\pm$ 0.30 & 74.41 $\pm$ 0.32 \\
        SiLU & 76.82 $\pm$ 0.24 & 73.19 $\pm$ 0.36 & 73.71 $\pm$ 0.39 & 75.59 $\pm$ 0.15 & 70.17 $\pm$ 0.35 & 73.49 $\pm$ 0.35 & 73.40 $\pm$ 0.21 \\
        Swish & 78.60 $\pm$ 0.28 & 75.69 $\pm$ 0.17 & 75.73 $\pm$ 0.19 & 78.18 $\pm$ 0.20 & 72.92 $\pm$ 0.18 & \textbf{77.97} $\pm$ 0.16 & 77.00 $\pm$ 0.35 \\
        Mish & 77.16 $\pm$ 0.24 & 73.58 $\pm$ 0.28 & 73.45 $\pm$ 0.20 & 75.74 $\pm$ 0.35 & 71.30 $\pm$ 0.09 & 73.39 $\pm$ 0.23 & 73.81 $\pm$ 0.27 \\
        SMU & 78.85 $\pm$ 0.19 & 75.36 $\pm$ 0.46 & 75.43 $\pm$ 0.20 & \textbf{78.28} $\pm$ 0.18 & 72.49 $\pm$ 0.19 & 77.84$\pm$ 0.16 & 76.89 $\pm$ 0.22 \\
        SMU-1 & 78.44 $\pm$ 0.26 & 76.01 $\pm$ 0.20 & 75.90 $\pm$ 0.23 & 76.71 $\pm$ 0.26 & 72.48 $\pm$ 0.11 & 77.46 $\pm$ 0.14 & 76.04 $\pm$ 0.10 \\ \midrule
        Swish-T & \textbf{79.02} $\pm$ 0.24 & 75.79 $\pm$ 0.23 & \textbf{76.04} $\pm$ 0.30 & 78.15 $\pm$ 0.29 & 72.76 $\pm$ 0.43 & 77.31 $\pm$ 0.14 & 77.00 $\pm$ 0.14 \\
        Swish-T$_{\text{B}}$ & 77.24 $\pm$ 0.99 & 75.84 $\pm$ 0.29 & 75.82 $\pm$ 0.29 & 77.80 $\pm$ 0.31 & \textbf{73.21} $\pm$ 0.19 & 77.40 $\pm$ 0.13 & 76.96 $\pm$ 0.12 \\
        Swish-T$_{\text{C}}$ & 78.72 $\pm$ 0.15 & \textbf{76.06} $\pm$ 0.36 & \textbf{76.04} $\pm$ 0.30 & 77.93 $\pm$ 0.20 & 72.75 $\pm$ 0.30 & 77.60 $\pm$ 0.26 & \textbf{77.15} $\pm$ 0.19 \\ \bottomrule
        \end{tabular}
        \end{adjustbox}
        \vspace{0.2cm}
        \caption{Comparison of the Swish-T family and various activation functions across different architectures on the CIFAR-100 dataset. The table shows the mean Top-1 accuracy and standard deviation of 5 runs.}
    \label{tab:cifar100}
    \end{table}

    \begin{figure}[h]
        \centering
        \begin{subfigure}[b]{0.49\textwidth}
            \centering
            \includegraphics[width=\textwidth]{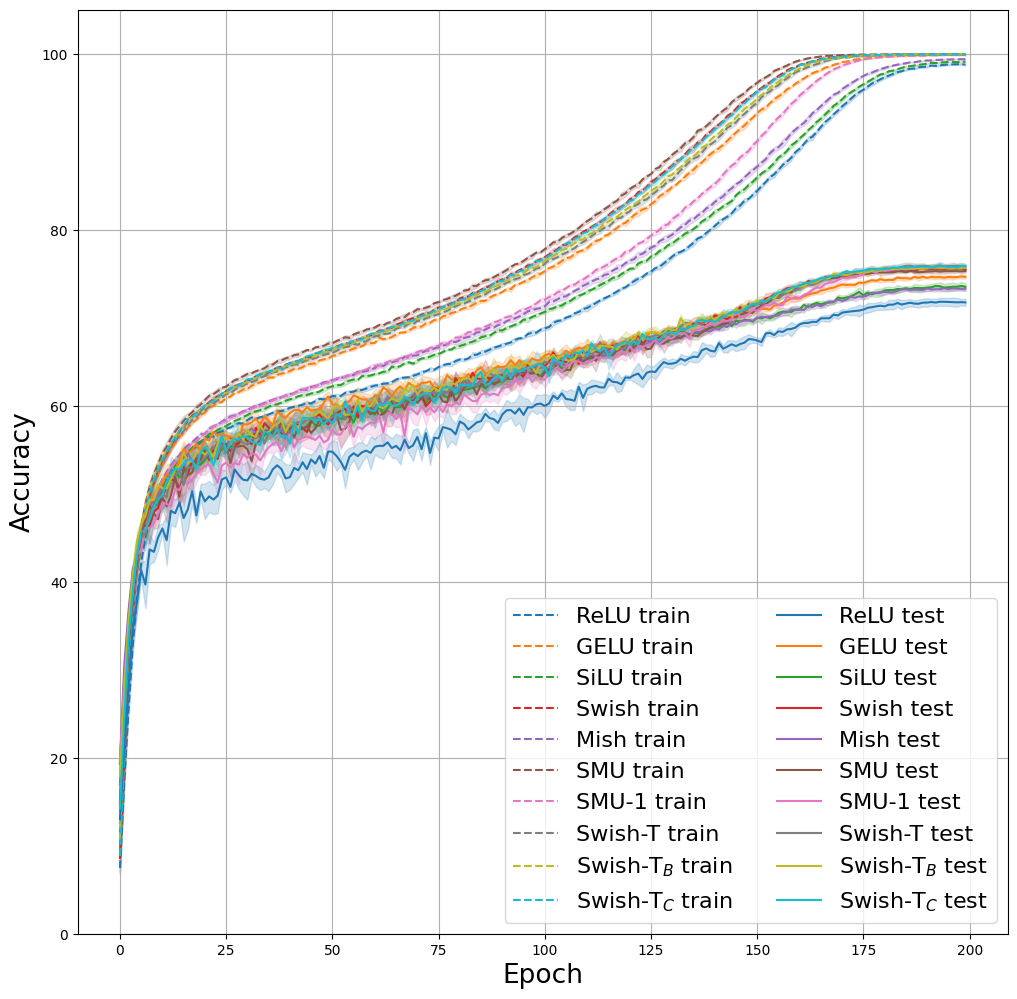}
            \caption{Train and test accuracy}
            \label{fig:shufflev2_acc}
        \end{subfigure}
        \hfill
        \begin{subfigure}[b]{0.49\textwidth}
            \centering
            \includegraphics[width=\textwidth]{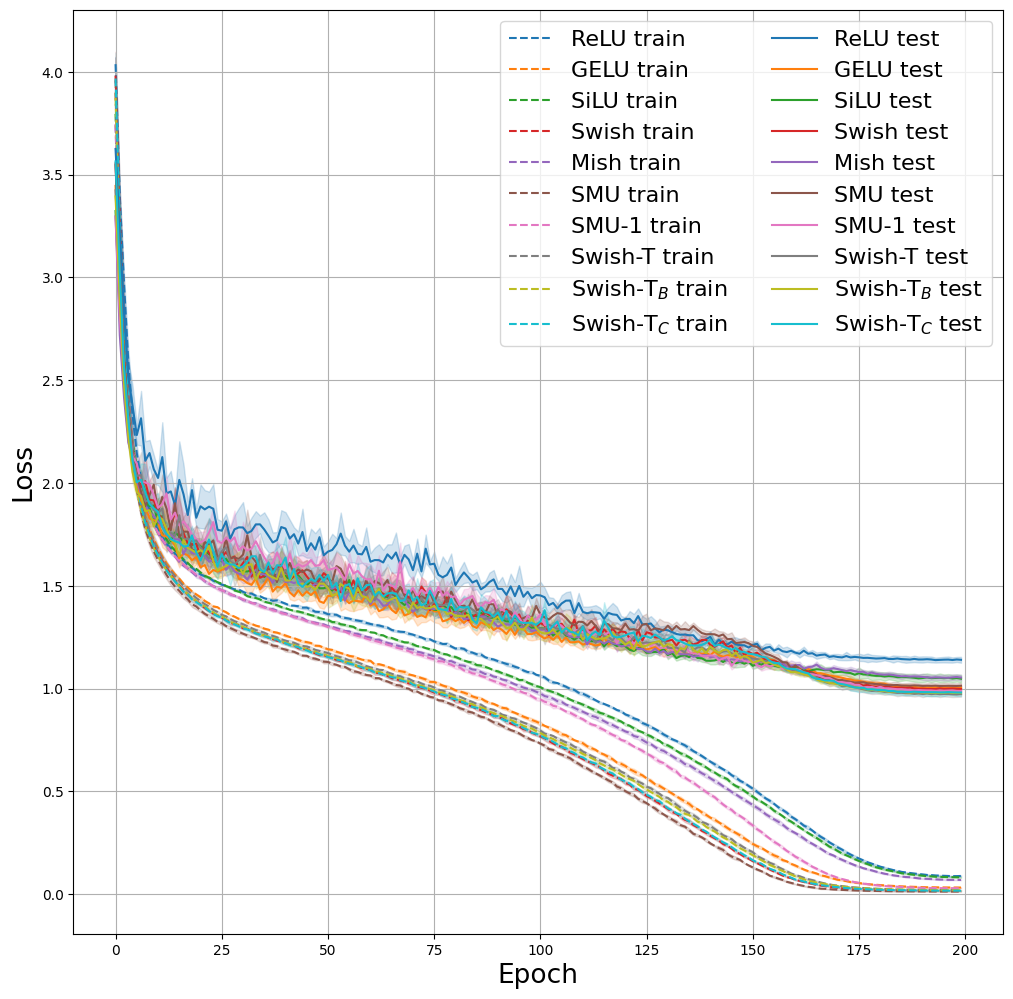}
            \caption{Train and test loss}
            \label{fig:shuffv2_loss}
        \end{subfigure}
        \vspace{0.2cm}
        \caption{Train and test curves for ShuffleNetv2 (2.x) on the CIFAR100 dataset. This figure shows the comparison of the performance metrics (Top-1 accuracy and loss) between the Swish-T family and other activation functions. The shaded areas represent the standard deviation.}
        \label{fig:curve_shuffv2}
    \end{figure}
    
    \begin{figure}[t]
        \centering
        \begin{subfigure}[b]{0.49\textwidth}
            \centering
            \includegraphics[width=\textwidth]{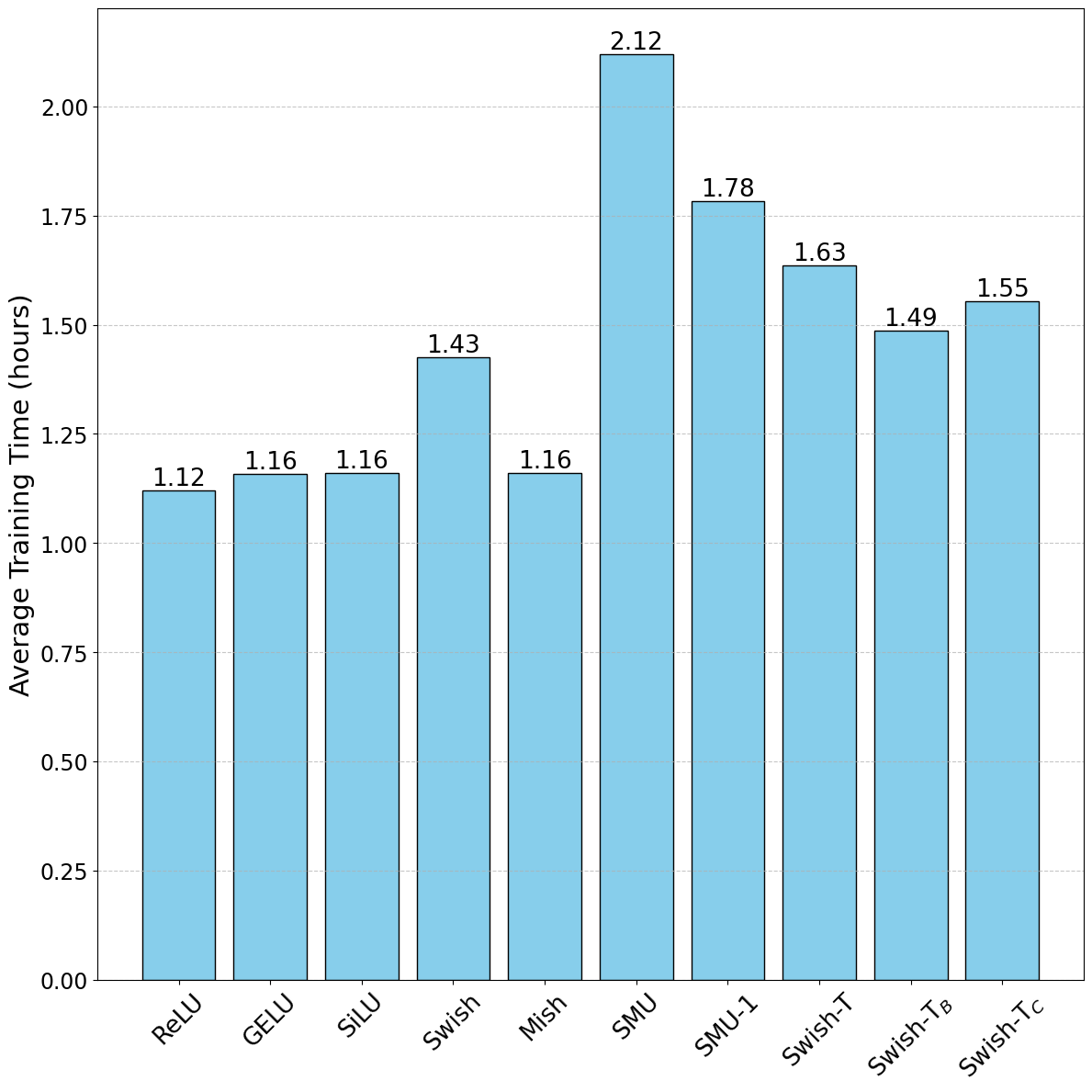}
            \caption{SENet-18}
            \label{fig:time_se}
        \end{subfigure}
        \hfill
        \begin{subfigure}[b]{0.49\textwidth}
            \centering
            \includegraphics[width=\textwidth]{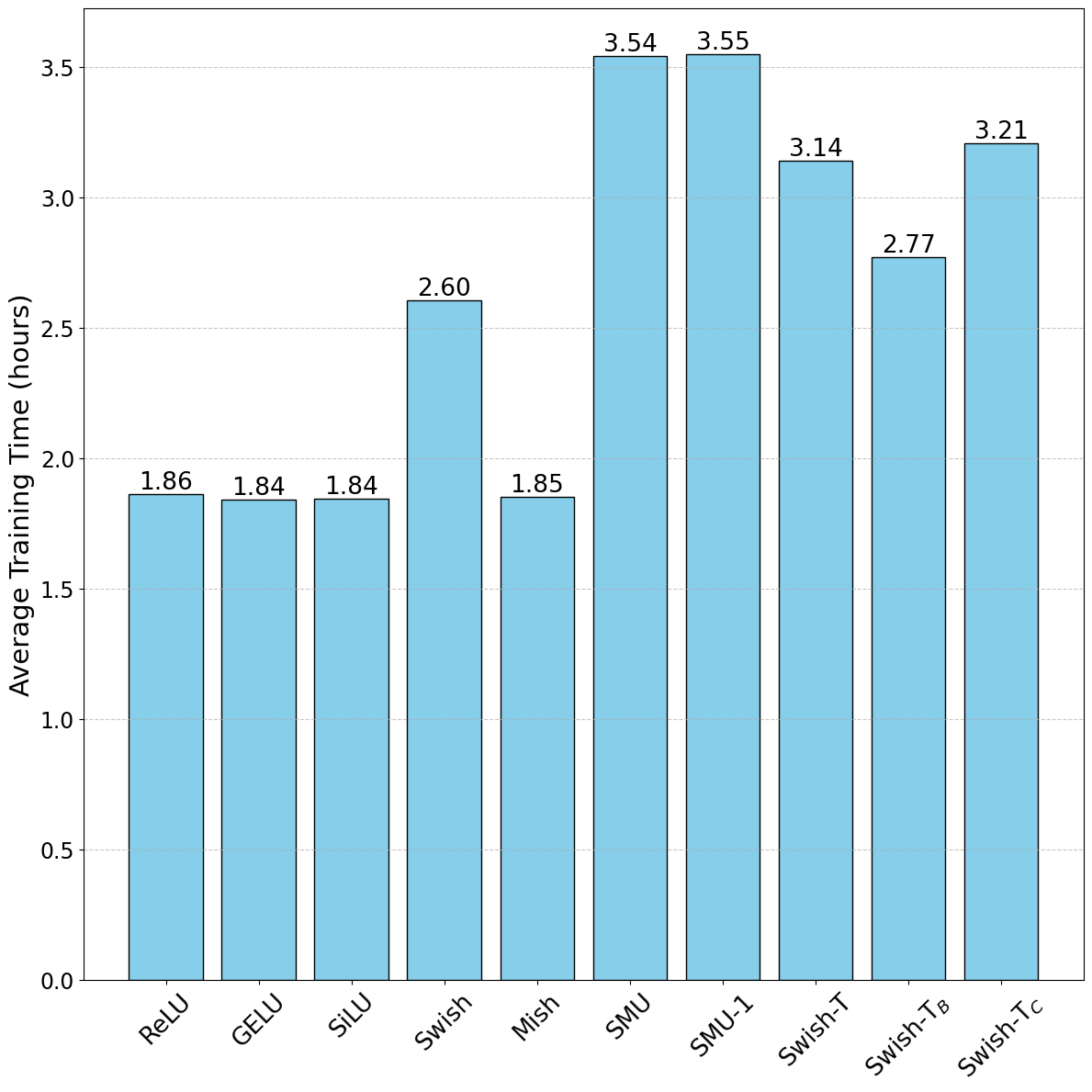}
            \caption{DenseNet-121}
            \label{fig:time_dense}
        \end{subfigure}
        \vspace{0.2cm}
        \caption{Average training time for SENet-18 and DenseNet-121 on CIFAR-10 using a single GPU. (Performance metrics can be found in Table \ref{tab:cifar10}.)}
        \label{fig:time_10}
    \end{figure}

    Tables~\ref{tab:cifar10} and~\ref{tab:cifar100} demonstrate the performance of different activation functions on CIFAR-10 and CIFAR-100 datasets across various deep learning architectures. The Swish-T family of functions consistently outperforms traditional activation functions like ReLU and GELU. Notably, Swish-T$_{\textbf{C}}$ achieves the highest accuracy of 94.27\% with the ShuffleNetV2 (2.x) model on CIFAR-10, surpassing ReLU by 2.34\%. On the CIFAR-100 dataset, Swish-T$_{\textbf{C}}$ also shows superior performance, with an improvement of 4.12\% over ReLU. Fig.~\ref{fig:curve_shuffv2} illustrates the train and test curves for ShuffleNetV2 (2.x) on CIFAR-100, providing insights into the effect of various activation functions, including the Swish-T family, on accuracy and loss over training epochs.

    In DenseNet-121, both SMU and Swish-T$_{\textbf{C}}$ reach the top accuracy of 94.95\% on CIFAR-10. However, as Fig.~\ref{fig:time_10} illustrates, Swish-T$_{\textbf{C}}$ exhibits faster training times compared to SMU. These results suggest that Swish-T variants not only enhance accuracy across diverse models and datasets but also offer computational efficiency. Fig~\ref{fig:time_10} further indicates that the Swish-T family maintains comparable or faster training times compared to SMU and SMU-1, making them effective choices for practical applications where both performance and speed are crucial.

\section{Ablation Study}

    We run an ablation to analyze the proposed Swish-T$_{\text{B}}$ and Swish-T$_{\text{C}}$ activations. In our study, these Swish-T variants are utilized without trainable parameters, specifically by fixing the parameter \( \beta \) instead of updating it during training. This approach allows us to achieve higher performance and faster training speed in some models. The fixed \( \beta \) simplifies the activation function, maintaining its effectiveness while reducing the complexity involved in training.
    
\subsection{Effect of Beta Fixation on ResNet-18 with Swish-T Family}

    Our experimental results, as shown in Tables ~\ref{tab:cifar10} and \ref{tab:cifar100}, indicate that the Swish-T family converges at similar beta values in some architectures. As demonstrated in Table~\ref{tab:beta_values}, Swish-T$_{\textbf{C}}$ converges with beta values ranging from 6.x to 7.x, and both Swish and the Swish-T family also converge in a similar range. However, in ResNet-18, Swish-T$_{\textbf{C}}$ and Swish-T$_{\textbf{B}}$ converge with beta values around 1.x and do not exhibit the highest performance among the compared activation functions. We hypothesized that the simple ResNet model, composed solely of skip-connections, makes it difficult for beta to reach an optimal point during training. Therefore, we arbitrarily set beta to 6.0, which we considered the optimal point based on Table \ref{tab:beta_values}. In Swish-T$_{\textbf{C}-6}$ and Swish-T$_{\textbf{B}-6}$, beta is fixed at 6.0 and not updated during training. When we trained ResNet-18 with Swish-T$_{\textbf{C}-6}$ and Swish-T$_{\textbf{B}-6}$, the results, as shown in Table~\ref{tab:ab_results_100}, demonstrated improved performance compared to previous experiments and higher performance than those shown in Tables~\ref{tab:cifar10} and \ref{tab:cifar100}.

    \begin{table}[h]
        \centering
        \begin{adjustbox}{max width=\textwidth}
            \begin{tabular}{c c c c c c c c}
            \hline
            \textbf{Activation Function} & \textbf{RN-18} & \textbf{SF-V2 (1.x)} & \textbf{SF-V2 (2.x)} & \textbf{SENet-18} & \textbf{EF-B0} & \textbf{MN-V2} & \textbf{DN-121} \\ \hline
            \multicolumn{8}{c}{\textbf{Beta Value}} \\ \hline
            Swish & 6.43 $\pm$ 0.02 & 6.57 $\pm$ 0.03 & 6.57 $\pm$ 0.02 & 5.96 $\pm$ 0.01 & 6.97 $\pm$ 0.02 & 7.22 $\pm$ 0.10 & 6.86 $\pm$ 0.04 \\ 
            Swish-T & 6.45 $\pm$ 0.02 & 6.77 $\pm$ 0.04 & 6.76 $\pm$ 0.04 & 5.94 $\pm$ 0.03 & 7.50 $\pm$ 0.02 & 7.56 $\pm$ 0.03 & 7.11 $\pm$ 0.06 \\ 
            Swish-T$_{\text{B}}$ & 1.49 $\pm$ 0.02 & 6.74 $\pm$ 0.05 & 6.81 $\pm$ 0.03 & 6.07 $\pm$ 0.01 & 7.40 $\pm$ 0.20 & 7.33 $\pm$ 0.02 & 7.19 $\pm$ 0.02 \\ 
            Swish-T$_{\text{C}}$ & 1.52 $\pm$ 0.10 & 6.77 $\pm$ 0.03  & 6.76 $\pm$ 0.04 & 6.16 $\pm$ 0.01 & 7.30 $\pm$ 0.03 & 7.38 $\pm$ 0.03 & 7.14 $\pm$ 0.02 \\ \hline
            \end{tabular}
        \end{adjustbox}
        \vspace{0.2cm}
        \caption{Comparison of beta values and standard deviations across different models trained on CIFAR-100.}
        \label{tab:beta_values}
    \end{table}
    
    These results indicate that the Swish-T family can be used in a non-parametric manner without training parameters, and by omitting the beta update process, it can achieve faster and similar or higher performance compared to parametric functions.
    
    \begin{table}[h]
        \centering
        \begin{adjustbox}{max width=\textwidth}
            \hfill
            \begin{tabular}{|c|c|c|}
            \hline
            \multicolumn{3}{|c|}{\textbf{CIFAR-10}} \\ \hline
            \textbf{Activation Function} & \textbf{Beta Value} & \textbf{Top-1} \\ \hline
            Swish & 5.81 $\pm$ 0.12 & 95.49 $\pm$ 0.07 \\ \hline
            Swish-T$_{\textbf{B}}$ & 1.49 $\pm$ 0.09 & 95.17 $\pm$ 0.21 \\ \hline
            Swish-T$_{\textbf{C}}$ & 1.43 $\pm$ 0.05 & 95.29 $\pm$ 0.06 \\ \hline \hline
            Swish-T$_{\textbf{B}-6}$ & 6.0 (\text{fiexd}) & 95.45 $\pm$ 0.14 \\ \hline
            Swish-T$_{\textbf{C}-6}$ & 6.0 (\text{fiexd}) & \textbf{95.54} $\pm$ 0.22 \\ \hline
            \end{tabular}
            \hspace{0.5cm}
            \begin{tabular}{|c|c|c|c|}
            \hline
            \multicolumn{4}{|c|}{\textbf{CIFAR-100}} \\ \hline
            \textbf{Activation Function} & \textbf{Beta Value} & \textbf{Top-1} & \textbf{Top-5} \\ \hline
            Swish & 6.43 $\pm$ 0.02 & 78.60 $\pm$ 0.28 & 94.18 $\pm$ 0.14 \\ \hline
            Swish-T$_{\textbf{B}}$ & 1.49 $\pm$ 0.02 & 77.23 $\pm$ 0.99 & 93.18 $\pm$ 0.50 \\ \hline
            Swish-T$_{\textbf{C}}$ & 1.52 $\pm$ 0.10 & 78.72 $\pm$ 0.15 & 94.10 $\pm$ 0.25 \\ \hline \hline
            Swish-T$_{\textbf{B}-6}$ & 6.0 (\text{fiexd}) & 78.81 $\pm$ 0.27 & 94.10 $\pm$ 0.06 \\ \hline
            Swish-T$_{\textbf{C}-6}$ & 6.0 (\text{fiexd}) & \textbf{79.06} $\pm$ 0.04 & \textbf{94.29} $\pm$ 0.07 \\ \hline
            \end{tabular}
        \end{adjustbox}
        \vspace{0.2cm}
        \caption{Performance Comparison of ResNet-18 with Different Activation Functions on CIFAR-10 and CIFAR-100. 
        }    
        \label{tab:ab_results_100}
    \end{table}

\section{Conclusion}

    In this work, we proposed the Swish-T family, which improves the Swish activation function by introducing an adaptive bias based on the input value \(x\). Swish-T combines a Tanh bias, offering superior performance but slower learning speeds. Swish-T$_{\textbf{A}}$ simplifies the formula for faster learning, while Swish-T$_{\textbf{B}}$ reintroduces the \(\beta\) parameter for better performance metrics. Swish-T$_{\textbf{C}}$ effectively controls the bias using \(\beta\), achieving stable performance.
    
    Our experimental results demonstrate that the Swish-T family outperforms traditional activation functions in various deep learning models, with Swish-T$_{\textbf{C}}$ showing the best overall performance. Through ablation studies, we confirmed that Swish-T$_{\textbf{B}}$ and Swish-T$_{\textbf{C}}$ exhibit high performance even when used as non-parametric functions with magic numbers for \(\beta\). Furthermore, they benefit from fast learning speeds and low memory usage since they do not have trainable parameters. The Swish-T family maintains zero-centering characteristics and controls the receptive field in the negative domain, making it a robust and versatile choice for diverse neural network architectures. Future research could further optimize these functions for specific applications.

\bibliographystyle{unsrt}
\bibliography{references}

\begin{thebibliography}{10}

\bibitem{hahnloser2000digital}
Richard~HR Hahnloser, Rahul Sarpeshkar, Misha~A Mahowald, Rodney~J Douglas, and H~Sebastian Seung.
\newblock \href{https://www.nature.com/articles/35016072}{Digital selection and analogue amplification coexist in a cortex-inspired silicon circuit}.
\newblock {\em Nature}, 405(6789):947--951, 2000.

\bibitem{jarrett2009best}
Kevin Jarrett, Koray Kavukcuoglu, Marc'Aurelio Ranzato, and Yann LeCun.
\newblock \href{https://ieeexplore.ieee.org/document/5459469}{What is the best multi-stage architecture for object recognition?}
\newblock In {\em 2009 IEEE 12th international conference on computer vision}, pages 2146--2153. IEEE, 2009.

\bibitem{nair2010rectified}
Vinod Nair and Geoffrey~E Hinton.
\newblock \href{https://dl.acm.org/doi/10.5555/3104322.3104425}{Rectified linear units improve restricted boltzmann machines}.
\newblock In {\em Proceedings of the 27th international conference on machine learning (ICML-10)}, pages 807--814, 2010.

\bibitem{lrelu}
Andrew~L. Maas, Awni~Y. Hannun, and Andrew~Y. Ng.
\newblock \href{https://ai.stanford.edu/~amaas/papers/relu_hybrid_icml2013_final.pdf}{Rectifier nonlinearities improve neural network acoustic models}.
\newblock In {\em in ICML Workshop on Deep Learning for Audio, Speech and Language Processing}, 2013.

\bibitem{he2015delving}
Kaiming He, Xiangyu Zhang, Shaoqing Ren, and Jian Sun.
\newblock \href{https://ieeexplore.ieee.org/document/7410480}{Delving Deep into Rectifiers: Surpassing Human-Level Performance on ImageNet Classification}, 2015.

\bibitem{hendrycks2023gaussian}
Dan Hendrycks and Kevin Gimpel.
\newblock \href{https://arxiv.org/abs/1606.08415}{Gaussian Error Linear Units (GELUs)}, 2023.

\bibitem{ramachandran2017searching}
Prajit Ramachandran, Barret Zoph, and Quoc~V. Le.
\newblock \href{https://arxiv.org/abs/1710.05941}{Searching for Activation Functions}, 2017.

\bibitem{Ma_2021}
Ningning Ma, Xiangyu Zhang, Ming Liu, and Jian Sun.
\newblock \href{https://ieeexplore.ieee.org/document/9577874}{Activate or Not: Learning Customized Activation}.
\newblock In {\em 2021 IEEE/CVF Conference on Computer Vision and Pattern Recognition (CVPR)}. IEEE, June 2021.

\bibitem{Biswas_2022}
Koushik Biswas, Sandeep Kumar, Shilpak Banerjee, and Ashish~Kumar Pandey.
\newblock \href{https://cdn.aaai.org/ojs/20557/20557-13-24570-1-2-20220628.pdf}{ErfAct and Pserf: Non-monotonic Smooth Trainable Activation Functions}.
\newblock {\em Proceedings of the AAAI Conference on Artificial Intelligence}, 36(6):6097–6105, June 2022.

\bibitem{misra2020mish}
Diganta Misra.
\newblock \href{https://www.bmvc2020-conference.com/assets/papers/0928.pdf}{Mish: A Self Regularized Non-Monotonic Activation Function}, 2020.

\bibitem{9878772}
Koushik Biswas, Sandeep Kumar, Shilpak Banerjee, and Ashish~Kumar Pandey.
\newblock \href{https://ieeexplore.ieee.org/document/9878772}{Smooth Maximum Unit: Smooth Activation Function for Deep Networks using Smoothing Maximum Technique}.
\newblock In {\em 2022 IEEE/CVF Conference on Computer Vision and Pattern Recognition (CVPR)}, pages 784--793, 2022.

\bibitem{goodfellow2013maxout}
Ian~J. Goodfellow, David Warde-Farley, Mehdi Mirza, Aaron Courville, and Yoshua Bengio.
\newblock \href{https://arxiv.org/abs/1302.4389}{Maxout Networks}, 2013.

\bibitem{alcaide2018eswish}
Eric Alcaide.
\newblock \href{https://arxiv.org/abs/1801.07145}{E-swish: Adjusting Activations to Different Network Depths}, 2018.

\bibitem{9301059}
Marina~Adriana Mercioni and Stefan Holban.
\newblock \href{https://ieeexplore.ieee.org/document/9301059}{P-Swish: Activation Function with Learnable Parameters Based on Swish Activation Function in Deep Learning}.
\newblock In {\em 2020 International Symposium on Electronics and Telecommunications (ISETC)}, pages 1--4, 2020.

\bibitem{9465622}
Marina~Adriana Mercioni and Stefan Holban.
\newblock \href{https://ieeexplore.ieee.org/document/9465622}{Soft-Clipping Swish: A Novel Activation Function for Deep Learning}.
\newblock In {\em 2021 IEEE 15th International Symposium on Applied Computational Intelligence and Informatics (SACI)}, pages 225--230, 2021.

\bibitem{math7121170}
Natinai Jinsakul, Cheng-Fa Tsai, Chia-En Tsai, and Pensee Wu.
\newblock \href{https://www.mdpi.com/2227-7390/7/12/1170}{Enhancement of Deep Learning in Image Classification Performance Using Xception with the Swish Activation Function for Colorectal Polyp Preliminary Screening}.
\newblock {\em Mathematics}, 7(12), 2019.

\bibitem{Tan_2020}
Mingxing Tan, Ruoming Pang, and Quoc~V. Le.
\newblock \href{https://ieeexplore.ieee.org/document/9156454}{EfficientDet: Scalable and Efficient Object Detection}.
\newblock In {\em 2020 IEEE/CVF Conference on Computer Vision and Pattern Recognition (CVPR)}. IEEE, June 2020.

\bibitem{farheen2020skin}
Misba Farheen, M~Manjushree, and Manish~Kumar Pandit.
\newblock \href{https://www.ijert.org/research/skin-cancer-detection-using-cnn-with-swish-activation-function-IJERTCONV8IS14022.pdf}{Skin Cancer Detection using CNN with Swish Activation Function}.
\newblock 2020.

\bibitem{Eger_2018}
Steffen Eger, Paul Youssef, and Iryna Gurevych.
\newblock \href{https://aclanthology.org/D18-1472/}{Is it Time to Swish? Comparing Deep Learning Activation Functions Across NLP tasks}.
\newblock In {\em Proceedings of the 2018 Conference on Empirical Methods in Natural Language Processing}. Association for Computational Linguistics, 2018.

\bibitem{pmlr-v97-wang19p}
Shengjie Wang, Tianyi Zhou, and Jeff Bilmes.
\newblock \href{https://proceedings.mlr.press/v97/wang19p.html}{Bias Also Matters: Bias Attribution for Deep Neural Network Explanation}.
\newblock In Kamalika Chaudhuri and Ruslan Salakhutdinov, editors, {\em Proceedings of the 36th International Conference on Machine Learning}, volume~97 of {\em Proceedings of Machine Learning Research}, pages 6659--6667. PMLR, 09--15 Jun 2019.

\bibitem{cifar}
Alex Krizhevsky, Geoffrey Hinton, et~al.
\newblock \href{https://www.cs.toronto.edu/~kriz/cifar.html}{Learning multiple layers of features from tiny images}.
\newblock 2009.

\bibitem{resnet}
Kaiming He, Xiangyu Zhang, Shaoqing Ren, and Jian Sun.
\newblock \href{https://ieeexplore.ieee.org/stamp/stamp.jsp?arnumber=7780459}{Deep Residual Learning for Image Recognition}, 2015.

\bibitem{tensorflow}
Mart\'{\i}n Abadi, Ashish Agarwal, Paul Barham, Eugene Brevdo, Zhifeng Chen, Craig Citro, Greg~S. Corrado, Andy Davis, Jeffrey Dean, Matthieu Devin, Sanjay Ghemawat, Ian Goodfellow, Andrew Harp, Geoffrey Irving, Michael Isard, Yangqing Jia, Rafal Jozefowicz, Lukasz Kaiser, Manjunath Kudlur, Josh Levenberg, Dandelion Man\'{e}, Rajat Monga, Sherry Moore, Derek Murray, Chris Olah, Mike Schuster, Jonathon Shlens, Benoit Steiner, Ilya Sutskever, Kunal Talwar, Paul Tucker, Vincent Vanhoucke, Vijay Vasudevan, Fernanda Vi\'{e}gas, Oriol Vinyals, Pete Warden, Martin Wattenberg, Martin Wicke, Yuan Yu, and Xiaoqiang Zheng.
\newblock \href{https://www.tensorflow.org/}{{TensorFlow}: Large-Scale Machine Learning on Heterogeneous Systems}, 2015.
\newblock Software available from tensorflow.org.

\bibitem{pytorch}
Adam Paszke, Sam Gross, Francisco Massa, Adam Lerer, James Bradbury, Gregory Chanan, Trevor Killeen, Zeming Lin, Natalia Gimelshein, Luca Antiga, Alban Desmaison, Andreas Köpf, Edward Yang, Zach DeVito, Martin Raison, Alykhan Tejani, Sasank Chilamkurthy, Benoit Steiner, Lu~Fang, Junjie Bai, and Soumith Chintala.
\newblock \href{https://pytorch.org/}{PyTorch: An Imperative Style, High-Performance Deep Learning Library}, 2019.

\bibitem{backpropagation1974}
Paul Werbos and Paul John.
\newblock \href{https://www.researchgate.net/publication/35657389_Beyond_regression_new_tools_for_prediction_and_analysis_in_the_behavioral_sciences}{Beyond regression : new tools for prediction and analysis in the behavioral sciences /}.
\newblock 01 1974.

\bibitem{Elfwing_2018}
Stefan Elfwing, Eiji Uchibe, and Kenji Doya.
\newblock \href{https://doi.org/10.1016/j.neunet.2017.12.012}{Sigmoid-weighted linear units for neural network function approximation in reinforcement learning}.
\newblock {\em Neural Networks}, 107:3–11, November 2018.

\bibitem{backpropagation1988}
Y.~LeCun, B.~Boser, J.~S. Denker, D.~Henderson, R.~E. Howard, W.~Hubbard, and L.~D. Jackel.
\newblock \href{https://ieeexplore.ieee.org/document/6795724}{Backpropagation Applied to Handwritten Zip Code Recognition}.
\newblock {\em Neural Computation}, 1(4):541--551, 1989.

\bibitem{mnist}
Yann LeCun, Corinna Cortes, and CJ~Burges.
\newblock \href{http://yann.lecun.com/exdb/mnist/}{MNIST handwritten digit database}.
\newblock {\em ATT Labs [Online]. Available: http://yann.lecun.com/exdb/mnist}, 2, 2010.

\bibitem{fashion}
Han Xiao, Kashif Rasul, and Roland Vollgraf.
\newblock \href{https://github.com/zalandoresearch/fashion-mnist}{Fashion-mnist: a novel image dataset for benchmarking machine learning algorithms}.
\newblock {\em arXiv preprint arXiv:1708.07747}, 2017.

\bibitem{svhn}
Yuval Netzer, Tao Wang, Adam Coates, Alessandro Bissacco, Bo~Wu, and Andrew~Y Ng.
\newblock \href{http://ufldl.stanford.edu/housenumbers/}{Reading Digits in Natural Images with Unsupervised Feature Learning}.
\newblock 2011.

\bibitem{lenet}
Yann LeCun, Bernhard Boser, John Denker, Donnie Henderson, Richard Howard, Wayne Hubbard, and Lawrence Jackel.
\newblock \href{https://proceedings.neurips.cc/paper/1989/hash/53c3bce66e43be4f209556518c2fcb54-Abstract.html}{Handwritten digit recognition with a back-propagation network}.
\newblock {\em Advances in neural information processing systems}, 2, 1989.

\bibitem{sgd1}
H.~Robbins and S.~Monro.
\newblock \href{https://www.columbia.edu/~ww2040/8100F16/RM51.pdf}{A stochastic approximation method}.
\newblock {\em Annals of Mathematical Statistics}, 22:400--407, 1951.

\bibitem{sgd2}
J.~Kiefer and J.~Wolfowitz.
\newblock \href{https://doi.org/10.1214/aoms/1177729392}{Stochastic Estimation of the Maximum of a Regression Function}.
\newblock {\em Annals of Mathematical Statistics}, 23:462--466, 1952.

\bibitem{coslr}
Ilya Loshchilov and Frank Hutter.
\newblock \href{https://arxiv.org/abs/1608.03983}{Sgdr: Stochastic gradient descent with warm restarts}.
\newblock {\em arXiv preprint arXiv:1608.03983}, 2016.

\bibitem{shufflev2}
Ningning Ma, Xiangyu Zhang, Hai-Tao Zheng, and Jian Sun.
\newblock \href{https://openaccess.thecvf.com/content_ECCV_2018/html/Ningning_Light-weight_CNN_Architecture_ECCV_2018_paper.html}{Shufflenet v2: Practical guidelines for efficient cnn architecture design}.
\newblock In {\em Proceedings of the European conference on computer vision (ECCV)}, pages 116--131, 2018.

\bibitem{senet}
Jie Hu, Li~Shen, and Gang Sun.
\newblock Squeeze-and-excitation networks.
\newblock In {\em Proceedings of the IEEE conference on computer vision and pattern recognition}, pages 7132--7141, 2018.

\bibitem{efficientnet}
Mingxing Tan and Quoc Le.
\newblock \href{https://proceedings.mlr.press/v97/tan19a.html}{Efficientnet: Rethinking model scaling for convolutional neural networks}.
\newblock In {\em International conference on machine learning}, pages 6105--6114. PMLR, 2019.

\bibitem{mobilenetv2}
Mark Sandler, Andrew Howard, Menglong Zhu, Andrey Zhmoginov, and Liang-Chieh Chen.
\newblock \href{https://www.computer.org/csdl/proceedings-article/cvpr/2018/642000e510/17D45Wuc32W}{Mobilenetv2: Inverted residuals and linear bottlenecks}.
\newblock In {\em Proceedings of the IEEE conference on computer vision and pattern recognition}, pages 4510--4520, 2018.

\bibitem{densenet}
Gao Huang, Zhuang Liu, Laurens Van Der~Maaten, and Kilian~Q Weinberger.
\newblock \href{https://www.computer.org/csdl/proceedings-article/cvpr/2017/0457c261/12OmNBDQbld}{Densely connected convolutional networks}.
\newblock In {\em Proceedings of the IEEE conference on computer vision and pattern recognition}, pages 4700--4708, 2017.

\end{thebibliography}
\end{document}